\documentclass[letterpaper]{article} 
\usepackage{aaai25}  
\usepackage{times}  
\usepackage{helvet}  
\usepackage{courier}  
\usepackage[hyphens]{url}  
\usepackage{graphicx} 
\usepackage{natbib}  
\usepackage{caption} 
\usepackage{algorithm}
\usepackage{algorithmic}

\usepackage{amsmath}
\usepackage{amssymb}
\usepackage{mathtools}
\usepackage{amsthm}
\usepackage[utf8]{inputenc} 
\usepackage{multirow}
\usepackage{amsfonts}       
\usepackage{nicefrac}       
\usepackage{microtype}      
\usepackage{xcolor}         
\usepackage{subfigure}
\usepackage{subcaption}
\usepackage{enumitem}

\usepackage{newfloat}
\usepackage{listings}
\DeclareCaptionStyle{ruled}{labelfont=normalfont,labelsep=colon,strut=off} 
\floatstyle{ruled}
\newfloat{listing}{tb}{lst}{}
\floatname{listing}{Listing}

\title{DroughtSet: Understanding Drought Through Spatial-Temporal Learning}
\author {
   Xuwei Tan\textsuperscript{\rm 1},
    Qian Zhao\textsuperscript{\rm 2},
    Yanlan Liu\textsuperscript{\rm 2},
    Xueru Zhang\textsuperscript{\rm 1}
}
\affiliations {
    \textsuperscript{\rm 1}Department of Computer Science and Engineering, The Ohio State University\\
    \textsuperscript{\rm 2}School of Earth Sciences, The Ohio State University\\
    \{tan.1206, zhao.4243, liu.9367, zhang.12807\}@osu.edu
}

\begin{document}

\maketitle

\begin{abstract}
Drought is one of the most destructive and expensive natural disasters, severely impacting natural resources and risks by depleting water resources and diminishing agricultural yields. Under climate change, accurately predicting drought is critical for mitigating drought-induced risks. However, the intricate interplay among the physical and biological drivers that regulate droughts limits the predictability and understanding of drought, particularly at a subseasonal to seasonal (S2S) time scale. While deep learning has been demonstrated with potential in addressing climate forecasting challenges, its application to drought prediction has received relatively less attention. In this work, we propose a new dataset, DroughtSet, which integrates relevant predictive features and three drought indices from multiple remote sensing and reanalysis datasets across the contiguous United States (CONUS). DroughtSet specifically provides the machine learning community with a new real-world dataset to benchmark drought prediction models and more generally, time-series forecasting methods. Furthermore, we propose a spatial-temporal model \textit{SPDrought} to predict and interpret S2S droughts. Our model learns from the spatial and temporal information of physical and biological features to predict three types of droughts simultaneously. Multiple strategies are employed to quantify the importance of physical and biological features for drought prediction. Our results provide insights for researchers to better understand the predictability and sensitivity of drought to biological and physical conditions. We aim to contribute to the climate field by proposing a new tool to predict and understand the occurrence of droughts and provide the AI community with a new benchmark to study deep learning applications in climate science. Resources are available at https://github.com/osu-srml/DroughtSet.
\end{abstract}

\section{Introduction}

Drought is among the most disastrous and costly natural hazards, affecting water resources, agricultural yields, heat waves, and ecosystem carbon sink \cite{cook2018climate}. Drought typically occurs under precipitation deficit, which is frequently accompanied by abnormally high temperatures and low humidity, leading to high evapotranspiration rates that quickly deplete soil moisture. As the global temperatures continue to increase, droughts are setting in quicker and becoming more intense and more frequent \cite{trenberth2014global, tripathy2023climate}. In particular, recent studies have highlighted the increasing frequency of S2S droughts, which initiate and intensify quickly at time scales from weeks to months, impairing ecosystem functions and challenging drought risk management practices \cite{pendergrass2020flash, yuan2023global}. For example, precipitation deficits combined with record-high temperatures in 2012 led to rapid drought development across the central US within just two months, resulting in estimated losses exceeding \$30 billion.

Accurate prediction of droughts is crucial for societal preparedness and risk mitigation strategies \cite{otkin2018flash,white2017potential,white2022advances}, but it still remains a significant challenge. Existing drought prediction models include data-driven \citep{bonaccorso2015probabilistic, santos2014spring, wanders2015decadal}, physically-based \citep{wanders2016improved, wang2009multimodel}, and hybrid models \citep{wu2022dynamic}.  While some of these models consider the interaction of biological drivers, their representations are generally simplified. In addition, most existing drought predictive models focused on single drought type, e.g. meteorological drought, hydrological drought, or ecological drought, while neglecting their joint behaviors. Tackling these drought prediction challenges require a robust and interpretable data-driven method that systematically integrates datasets of relevant climate and vegetation features to jointly predict multiple aspects of drought \cite{aghakouchak2022status,hao2018seasonal}. However, most existing methods simplify or ignore dynamic interactions among different factors, which inhibits realizing the potential of artificial intelligence (AI) to improve drought prediction accuracy and advance mechanistic understanding of droughts. We discuss these in Appendix 6.2.

To address this research gap, we create a benchmark dataset, DroughtSet, by compiling climate, physical, and vegetation predictors that are relevant to drought initiation, development, and propagation from various remote sensing and reanalysis products across the contiguous United States (CONUS) during years 2003--2013. DroughtSet encompasses the wide diversity of climatic and ecological settings and includes frequent drought events in recent decades. Specifically, we collect and preprocess drought-related predictors as listed in Table \ref{tab:drought_datasets} (e.g. precipitation, temperature, elevation, leaf area index), consisting of both static variables and dynamic variables with coordinates. We also compile three drought indices: soil moisture drought measured by normalized surface soil moisture \cite{yuan2023global}, ecohydrological drought measured by the Evaporative Stress Index (ESI) \cite{otkin2014examining}, and ecological drought measured by solar-induced chlorophyll fluorescence (SIF) \cite{mohammadi2022flash}. Collectively, DroughtSet can be used to benchmark multivariate forecasting, spatiotemporal forecasting, and irregular forecasting (learning from static variables).  We hope to accelerate future research in drought predictions and benchmark deep learning-based methods by releasing this dataset.

In addition, we propose a multi-task SPatial-temporal framework for drought prediction on DroughtSet, referred to as \textit{SPDrought}, which exploits the spatial-temporal interconnections within and across climate and vegetation features. It accounts for influences of nearby regions by aggregating temporal features with neighboring locations and learns from both static and dynamic features to predict three drought indices. Furthermore, we employ the Integrated Gradient (IG) method, as described in \cite{sundararajan2017axiomatic}, to interpret and quantify how these features influence drought development across CONUS. The results will serve as a data-driven benchmark, informing further research towards enhancing the mechanistic understanding and simulation of S2S droughts in existing Earth system models. This, in turn, could potentially support the development of drought risk mitigation strategies under future climate. Our contributions are summarized as follows:
\begin{itemize}[leftmargin=*,itemsep=0cm]
\item We introduce DroughtSet, a drought prediction dataset for the machine learning community. It serves as a complementary resource to existing climate datasets.  DroughtSet is a collection of droughts indices and the corresponding climate, physical, and vegetation conditions, specifically focusing on the contiguous U.S. DroughtSet will be released upon the acceptance of this work.

\item To forecast drought, we propose \textit{SPDrought}, a spatial-temporal drought prediction model that incorporates geographic neighbor features fusion. It jointly uses both static and dynamic features to accurately predict three key drought indices.

\item We leverage the Integrated Gradient to interpret the hidden relationship between the predicted drought indices and the climate, physical, and vegetation features. The interpretability module offers new insights into the dependence structures of among the drought-related physical and ecological variables in the Earth system.
\end{itemize}

\section{DroughtSet}

In this section, we introduce DroughtSet, a collection of climate, physical, vegetation conditions, and drought indices from multiple publicly available remote sensing and reanalysis datasets. We selected these variables based on their relevance and potential influence on the mechanisms of drought initiation and development.

\subsection{Data Collection and Preprocessing} 

DroughtSet includes weekly climate-related data from 2003 to 2013~(11 years, 572 weeks) across CONUS covering an area of over 8 M $\text{km}^2$. The details of drought/feature types, variables, and their sources are outlined in Table \ref{tab:drought_datasets}.  To ensure consistency in geographical resolution, all variables have been resampled to a 4 km spatial scale. The spatial data are represented as a grid of $585\times1386$ pixels across CONUS, with the values in each pixel denoted as $P(i, j)$, where $(i, j)$ are the spatial coordinates of the pixel on the map.  Note that, 42\% of this pixel area consists of the ocean that is outside the scope here.  Only the remaining 58\% pixels are used in the analyses. Furthermore, all temporal variables are aggregated to a weekly time scale, detailed in Section \ref{sec:introduce_to_var}. For the static variables, both elevation and canopy height are numeric variables while land cover is a categorical variable with 97 categories. We also include the mean and standard deviation of the drought indices as the static variables.  In total, the dataset comprises $585\times1386\times 11 \times 52 \times 3$ Drought indices, $585\times1386\times 11 \times 52 \times 11$ dynamic predictors, and  $585\times1386 \times 9$ static predictors. 
Note that NaN values exist in the datasets due to different temporal coverages of remote sensing-derived products.

\begin{table*}[h!]
    \centering

    \caption{Variables used to quantify three types of droughts and their predictive features.}
    
\setlength{\tabcolsep}{1mm}
   \small
    \begin{tabular}{p{2.5cm} p{4.8cm} p{1.4cm} p{8cm}}
        \hline
        \textbf{Drought/Feature Type} & \multirow{2}{4.8cm}{\textbf{Variables}} & \textbf{Dynamic or Static} & \multirow{2}{8cm}{\textbf{Dataset \& Native Resolution}}\\


        \hline
 
Soil moisture \newline drought & \multirow{2}{4.8cm}{Soil Moisture across depths~(SM)} & \multirow{2}{2.5cm}{Dynamic} & \multirow{2}{8cm}{NLDAS \cite{xia2012continental}, hourly, 1/8° }
 \\

\hline

\multirow{3}{2.5cm}{Ecohydrological drought} & \multirow{3}{4.8cm}{Evaporative stress index (ESI)} & \multirow{3}{2.5cm}{Dynamic}  & {ALEXI \cite{holmes2018microwave, liu2021global}, weekly, 0.25° \newline GridMET \cite{abatzoglou2013development}, daily, 4 km}  \\

\hline    

Ecological \newline drought & \multirow{2}{4.8cm}{Solar-induced fluorescence (SIF)} & \multirow{2}{2.5cm}{Dynamic}  & \multirow{2}{8cm}{CSIF \cite{zhang2018global}, 4-day, 0.05°} \\

\hline
          
\multirow{3}{2.5cm}{Physical \& climate features} & 2m temperature, radiation, VPD, precipitation, wind speed, PET, PDSI, SP & \multirow{2}{1.5cm}{Dynamic} &   \multirow{2}{8cm}{GridMET \cite{abatzoglou2013development}, daily, 4 km \newline ERA5 \cite{munoz2021era5}, hourly, 9 km}\\ 
& elevation & Static &  SRTM \cite{nasa_jpl_2013}, 30 m \\

\hline

\multirow{4}{2.5cm}{Vegetation features} & VOD & Dynamic &  VODCA \cite{moesinger2020global}, daily, 0.25° \\
& LAI & Dynamic &  MODIS \cite{myneni2015mcd15a3h}, 8-day, 500 m \\
& Canopy Height & Static &  GLAD \cite{potapov2021mapping}, 30 m \\
& Land Cover & Static &  NLCD \cite{homer2012national,homer2020conterminous}, 30 m \\
        \hline
    \end{tabular}
    \label{tab:drought_datasets}
\end{table*}

\subsection{Drought indices and Predictors}
\label{sec:introduce_to_var}

DroughtSet focuses on three drought types: soil moisture drought measured by surface soil moisture (SM), ecohydrological drought measured by Evaporative Stress Index (ESI), and ecological drought measured by the solar-induced chlorophyll fluorescence (SIF), all normalized using their quantiles at each location.  These drought indices are denoted as $D_{i,j}(t) = \left[d_{i,j}^1(t),\cdots,d_{i,j}^K(t)\right]$, $K$ is the number of indices, which equals to 3 in this case.

\begin{itemize}[leftmargin=*,itemsep=0cm]
\item \textbf{Soil Moisture}: Low surface soil moisture reflects soil moisture drought intensity and is key to a wide range of ecosystem functions and drought propagation to other downstream drought types \cite{yuan2023global}.
\item  \textbf{Evaporative Stress Index:} ESI is the ratio between actual evapotranspiration and potential evapotranspiration. Low ESI values represent ecohydrological droughts with low moisture supply from the land relative to the demand of the atmosphere. ESI is affected by plant stomatal response to moisture deficit and regulates drought intensification through atmospheric feedback \citep{nguyen2019using}.
\item  \textbf{Solar-induced chlorophyll Fluorescence:} SIF is a surrogate highly correlated with gross primary productivity. Low SIF values reflect impaired photosynthetic activity, offering an effective representation of ecological droughts given its high sensitivity to water stresses  \citep{he2019impacts}.
\end{itemize}

\begin{figure}[hbtp]
    \centering
    \includegraphics[width=0.9\linewidth]{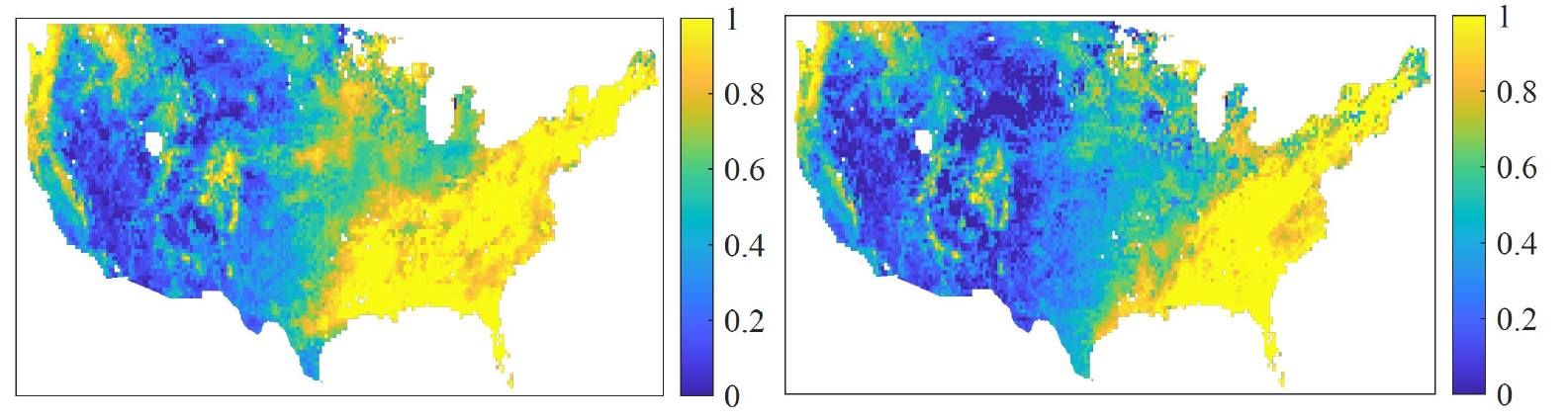}
    \caption{An example of drought development in July 2012. The left and right panels show the evaporative stress index in the 28th and 32nd weeks in 2012, respectively. ESI reduced in the Central Plains, indicating ecohydrological drought. }
    \label{fig:drought_demo}
\end{figure}

These metrics are of focus here for S2S drought because, unlike other commonly used drought severity indices such as the Standardized Precipitation Index and Palmer Drought Severity Index~(PDSI) that are typically used to capture interannual drought, these metrics have been demonstrated to respond quickly and have wide-ranging implications on water resources, ecosystem carbon sink strength, and agricultural productivity \citep{christian2021global,ford2023flash,he2019impacts,koster2019flash,nguyen2019using,yuan2023global}. In our work, these drought indices are used as the targets to be predicted.  Figure \ref{fig:drought_demo} visualizes the evaporative stress, using the 2012 central US drought as an example, which illustrates the pattern of a drought index.

To predict the three drought indices, we use attributes including climate and physical conditions, ecological conditions, and drought indices themselves in lagged time steps as features for drought prediction.  These features are categorized as temporal dynamic attributes and spatial static attributes that will be considered differently in \textit{SPDrought}. These features are described in the following.

\paragraph{Physical and climate conditions:} 
\begin{itemize}[leftmargin=*,itemsep=0cm]
\item \textbf{Elevation}: Elevation is a static numeric variable. It affects local climate,  subsurface hydrological processes, and thus drought occurrence and intensity.
\item \textbf{Air Temperature}: Air temperature and the following climate conditions are all dynamic numeric variables. High air temperatures accelerate soil moisture depletion and intensify atmospheric demand, thereby facilitating drought intensification.
\item \textbf{Precipitation}: It determines the amount of water input to the land. Precipitation deficit directly leads to drought. 
\item \textbf{Radiation}: Incident solar radiation controls the available energy that drives water loss from the land to the atmophere.  High radiation could accelerate water loss rate and thus drought onset. 
\item \textbf{Vapor Pressure Deficit~(VPD)}: VPD characterizes atmospheric moisture deficit and drives evapotranspiration through aerodynamic processes. 
\item \textbf{Wind Speed}: High wind speed enhances aerodynamic conductance and thus water and heat transfer rates from the land to the atmosphere, thereby potentially contributing to drought initiation development.
\item \textbf{Potential Evapotranspiration~(PET)}: PET represents the atmospheric water demand. 
\item \textbf{PDSI}: It quantifies the severity of meteorological drought based on meteorological conditions and an empirical water balance model, which primarily characterizes long-term droughts.
\item \textbf{Surface Pressure~(SP)}: SP is the atmospheric pressure at Earth's surface. Changes in surface pressure can regulate weather patterns, atmospheric moisture transport, and thus drought occurrence.

\item \textbf{SM Root}: Different from surface soil moisture, root zone soil moisture measures the amount of water available to plant water uptake.

\end{itemize}
\paragraph{Ecological conditions:} 
\begin{itemize}[leftmargin=*,itemsep=0cm]
\item \textbf{Biomass dynamics measured by Leaf Area Index (LAI)}: LAI represents the leaf area per ground unit area. We aggregate it from the original 8-day temporal steps to weekly steps using linear interpolation.
\item \textbf{Vegetation Optical Depth (VOD)}: VOD represents total vegetation water content.
\item \textbf{Canopy Height}:  Canopy height is a static categorical variable, representing ecosystem structure.
\item \textbf{Land Cover}: It is a static categorical variable, including categories such as forests, water bodies, and grasslands.
\end{itemize}

With these climate-related drought indices and attributes, each pixel has $N$ static features $S_{i,j} = \left[s_{i,j}^1,\cdots,s_{i,j}^N\right]$ (i.e., land cover, elevation, canopy height, long-term averages of drought indices and their standard deviations to capture variability) and $M$ dynamic features $X_{i,j}(t) = \left[x_{i,j}^1(t),\cdots,x_{i,j}^M(t)\right]$. The goal is to train a machine learning model $h\in\mathcal{H}$ from existing static and dynamic features $\left\{S_{i,j},D_{i,j}(t),X_{i,j}(t) \right\}_{i\in [I], j\in[J],t\in[T]}$ that can simultaneously predict multiple drought indices $\left\{D_{i,j}\left(T+\tau\right)\right\}_{\tau\geq 1}$ for any location $(i,j)$ in the future.  Because both drought indices $D_{i,j}(t)$ and dynamic features $X_{i,j}(t)$ are time-varying and jointly used for predictive tasks, we combine them and define $U_{i,j}(t) = \left[D_{i,j}(t),X_{i,j}(t)\right]$. 

In addition, as DroughtSet consists of both static features and temporal features for each location with geographic coordinates, it offers a versatile platform for benchmarking various forecasting methods.  This dataset can be utilized in univariate forecasting tasks, which focus on directly predicting drought indices from single variables.  It also supports multivariate forecasting, where multiple variables are used jointly to predict drought indices. Furthermore, DroughtSet is ideal for spatiotemporal forecasting, which leverages both spatial and temporal information to enhance prediction accuracy. Lastly, it can be employed in irregular forecasting tasks that jointly use static and temporal features, providing a comprehensive tool for advanced drought prediction models.

\begin{figure*}[h]
    \centering
    \includegraphics[width=0.8\linewidth]{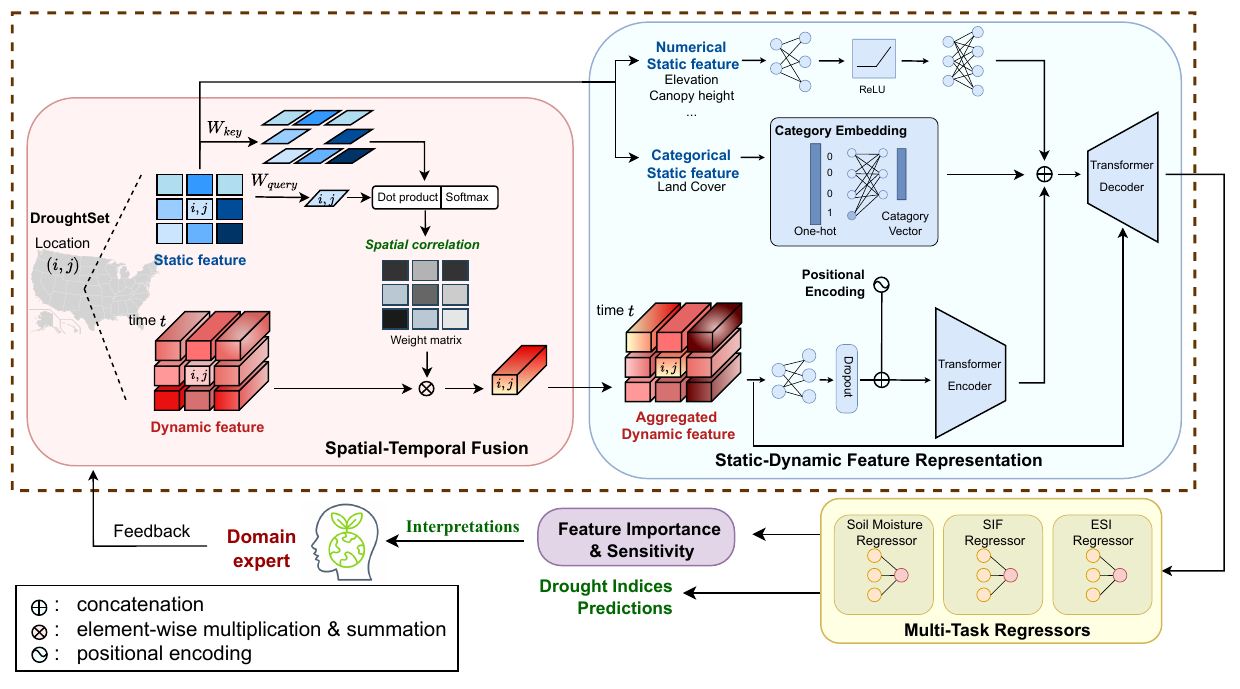}
    \caption{\textit{SPDrought} architecture for Forecasting Drought Indices: the spatial-temporal fusion module first exploits the spatial correlation of data with its neighbors using static features and leverages the learned correlation to aggregate the dynamic features; the static-dynamic feature representation exploits both spatial and temporal patterns with three network modules. Such representation is shared among multi-task regressors for generating multiple drought indices predictions. Subsequently, we analyze how individual features at various timestamps influence the final predictions using our interpretation method. Domain experts are encouraged to provide feedback on variable selection and model design, which can further refine the model and uncover deeper relationships among variables.}
    \label{fig:algo}
\end{figure*}

\section{Method}

Next, we introduce a comprehensive framework that utilizes both spatial and temporal information to predict drought indices. Our approach considers regional spatial similarity to aggregate information for robust prediction and introduces climate attribute-specific representation functions to learn from the hidden pattern of both static and time-series data.

\paragraph{Spatial-Temporal Fusion.} 
Since climate information from proximate geographical locations often
exhibits mutual influences, we hypothesize that data from neighboring locations may contain useful information that can help enhance the accuracy and reliability of the prediction. The key challenge is to exploit the spatial correlation and strategically leverage the learned correlation to enhance prediction at the target location. Intuitively, the target location may benefit more from those neighbors that are sufficiently correlated, e.g., sharing similar topography or land cover. Let neighborhood $ \mathcal{N}_{i,j} = \{(\Bar{i}, \Bar{j}) \mid \left(|\Bar{i}-i| \leq d, |\Bar{i}-j| \leq d , (\Bar{i}, \Bar{j}) \neq (i, j)\right) \}$, {where $d$ is a distance threshold}. Inspired by scaled dot-product attention mechanism \cite{vaswani2017attention}, we exploit the spatial correlation between any target location $(i, j)$ (known as a query in attention mechanism) and neighbors in $\mathcal{N}_{i,j}$ (known as keys) based on the following:
\begin{equation}
    A = \text{softmax}\left(\frac{(S_{i,j} W_{\text{query}})(S_{\mathcal{N}_{i,j}} W_{\text{key}})^T}{R_{i,j} \times \sqrt{N}}\right)
\end{equation}
where $S_{i,j}$ are $N$ static features at $(i, j)$, $S_{\mathcal{N}_{i,j}}$ is a matrix with each column the static features corresponding to one neighbor in $\mathcal{N}_{i,j}$. $W_{query} \in \mathbb{R}^{N \times N}$ and $W_{key} \in \mathbb{R}^{N \times N}$ are two linear transformation matrices that are learned to exploit spatial correlation. $R_{i,j}$ is a vector with each element the Euclidean distance between  $(i, j)$ and neighbors in $\mathcal{N}_{i,j}$, which leverages the prior spatial information to refine correlation learning process. For simplicity, we consider a $5\times5$ square area in this study, where $d$ is set to $2$ in this paper. To avoid division by zero, we manually set the distance to itself as 0.8. The spatial correlation weight $A$ can then be used to aggregate the regional time-varying attributes: $\Tilde{U}_{\Bar{i}, \Bar{j}}(t) = \Sigma_{\Bar{i}, \Bar{j} \in \mathcal{N}_{{(i, j)}}} A_{{\Bar{i}, \Bar{j}}} \cdot U_{\Bar{i}, \Bar{j}}(t)$.

\paragraph{Spatial-Temporal Representation and Multi-task Learning.}
Given $\{S_{i,j},\widetilde{U}_{i,j}(t)\}$, we next learn the representations of the climate data, which combine static and dynamic feature representations generated by separated networks:

\begin{itemize}[leftmargin=*]
    \item Static feature representation: Given a set of static features $S_{i,j}=\left[s_{i,j}^1,\cdots,s_{i,j}^N\right]$, we aim to obtain higher-level representation that encapsulates the underlying patterns among them. Because categorical (land cover type) and numerical (elevation, canopy height, long-term averages, and standard deviations of SM, SIF, and ESI) features have inherent differences in semantic meanings, we shall generate their representations differently. We apply two layers of MLP linked by the ReLU function to learn the representations of numerical features and adopt embedding approaches for categorical features to generate their representations, which we denote as $f^s_{i,j}$. 
    
    \item Dynamic feature representation: To learn the temporal patterns, especially the long-term dependencies of climate data, we first adapt \textit{Transformer} \cite{vaswani2017attention} encoder to generate temporal representations. Before the \textit{Transformer}, we expand the initial temporal features $\widetilde{U}_{i,j}(t)$  of our data (including $K$ drought indices and $M$ dynamic features with a total dimension of 14)  via linear transformation $W$ and project the dimensions to 48. This linear transformation also facilitates learning the interconnections among these distinct dynamic features. After integrating the \textit{positional encoding} $\mathrm{PE}(t)$, we generate temporal representation $f^t_{i,j}(t) = \mathrm{TransformerEncoder}\left(\widetilde{U}_{i,j}(t)\cdot W+\mathrm{PE}(t)\right)$.  We then concatenate the static representations to dynamic feature representations at each time stamp. The concatenated representations are fed  into the \textit{Transformer decoder} to generate representations $\{F_{i,j}(t')\}_{t'\in \{T+1...T+26\}}$ for the next 26 weeks.

\end{itemize}
 With the representation $\{F_{i,j}(t')\}_{t'\in \{T+1...T+26\}}$ for the prediction weeks, we employ three \textit{task-specific} regressors to map the representation of the next 26 weeks to drought indices. Specifically, let $\widehat{d}^k_{i,j}(t')=\mathrm{Regressor}_k\left(F_{i,j}(t')\right)$ be the prediction of $k$-th drought index $d^k_{i,j}(t')$ after $t'$ weeks. We jointly train all the parameters by minimizing the total loss between predictions $\widehat{d}^k_{i,j}(t')$ and ground-truth $d^k_{i,j}(t')$ for all drought indices at all locations (we use 
  a batch of locations to update the model at every iteration in implementation) under mean absolute error loss function $\mathcal{L}$:
\begin{equation}
\resizebox{.9\hsize}{!}{$\displaystyle \min \sum_{k\in [K]} \sum_{t'\in \{T+1...T+26\}}\sum_{i\in[I],j\in[J]}\mathcal{L}\left(\widehat{d}^k_{i,j}(t'),d^k_{i,j}(t')\right).$}
\label{eq:loss}
\end{equation}

\paragraph{Interpreting the impacts of attributes on drought.}

To identify which attributes and time steps contribute most to the final predictions,  we leverage {integrated gradient} \cite{sundararajan2017axiomatic} to investigate how their contributions to the final predictions change over time (i.e., how \textit{sensitive} the predictions are to these features). The integrated gradient estimates feature sensitivities by integrating the gradients of the model's output with respect to the input along a straight path from an ``input baseline" to the input. Then we quantify the importance of static variables by looking at the features that cause the larger performance change and are more important for prediction tasks.

\begin{table*}[h!]
\small
\centering
\caption{Average mean absolute error over three experiments (standard deviations are reported in the appendix).}
\label{tab:baselines}
\setlength{\tabcolsep}{1mm}
\begin{tabular}{ccccccccc}
\hline
MAE ($\times 10^{-3}$) & \textbf{SPDrought} & \textbf{Transformer}  & \textbf{Informer}  & \textbf{PatchTST} & \textbf{DLinear} & \textbf{iTransformer} & \textbf{TimesNet} & \textbf{LSTM} \\
\hline
\textbf{Soil Moisture}            & \textbf{21.39}  & 34.56  & 38.08  & 36.32  & 47.61  & 32.34  & 25.96  & 31.36  \\
\textbf{ESI} & \textbf{4.40}   & 5.99   & 6.37   & 6.37   & 6.82   & 6.06   & 5.11   & 5.83   \\
\textbf{SIF}                      & \textbf{12.21}  & 16.00  & 17.71  & 21.36  & 20.99  & 15.47  & 14.11  & 15.35  \\
\textbf{Total}                    & \textbf{38.01}  & 56.56  & 62.16  & 64.05  & 75.41  & 53.87  & 45.18  & 52.54  \\
\hline
\end{tabular}
\end{table*}

\begin{figure*}[h]
   \centering
   \subfigure[Influence of Surface Pressure]{\includegraphics[width=0.28\linewidth]{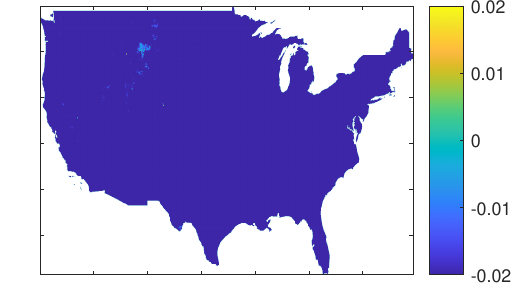}}
   \subfigure[Influence of Radiation]
   {\includegraphics[width=0.28\linewidth]{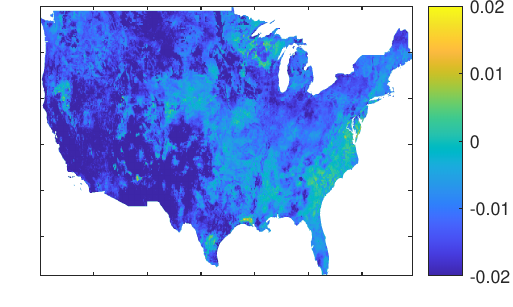}}
   \subfigure[Influence of PET]
   {\includegraphics[width=0.28\linewidth]{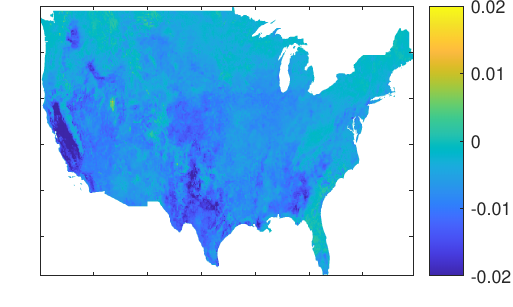}}
   \caption{The sensitivity of soil moisture to the top three predictive features measured by the integrated gradient, including surface pressure, radiation, and PET. }
    \label{fig:intersm}
\end{figure*}
\begin{figure*}[h]
    \centering
       \subfigure[Influence of Radiation]{\includegraphics[width=0.28\linewidth]{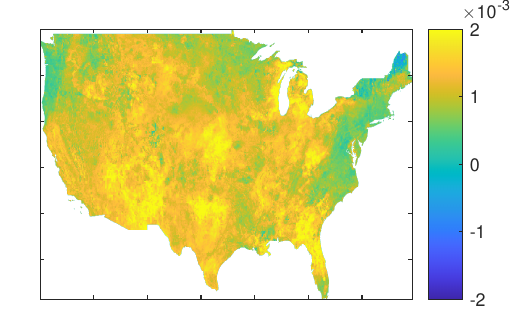}}
       \subfigure[Influence of Pressure]{\includegraphics[width=0.28\linewidth]{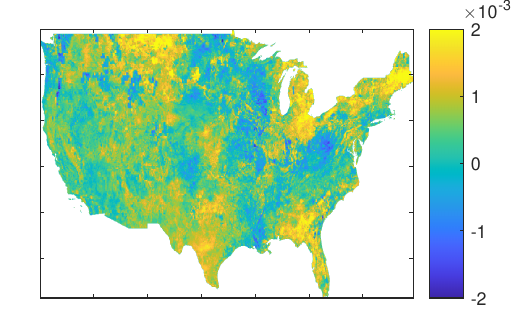}}
       \subfigure[Influence of SIF]{\includegraphics[width=0.28\linewidth]{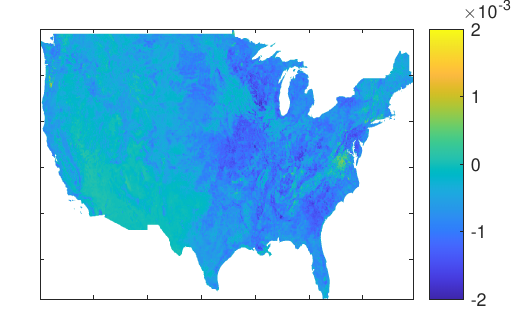}}
    \caption{The sensitivity of Evaporative Stress Index. Radiation and Pressure show a positive influence on the evaporative stress index while SIF reflects a minor negative influence.  }
    \label{fig:interesi}
\end{figure*}
\begin{figure*}[h]
    \centering
       \subfigure[Influence of Pressure]{\includegraphics[width=0.28\linewidth]{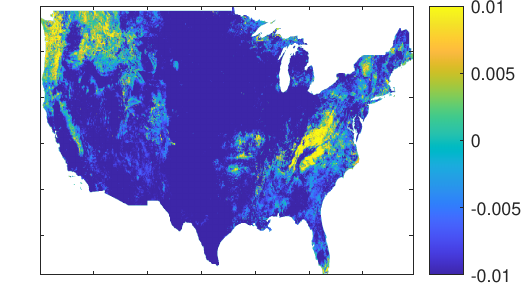}}
       \subfigure[Influence of Radiation]{\includegraphics[width=0.28\linewidth]{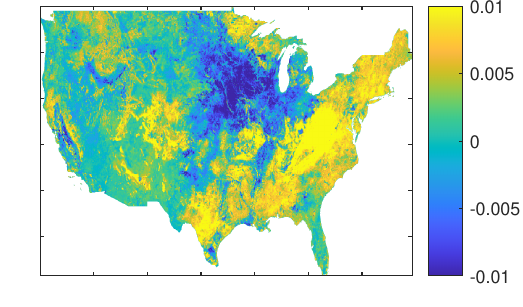}}
       \subfigure[Influence of SM root]{\includegraphics[width=0.28\linewidth]{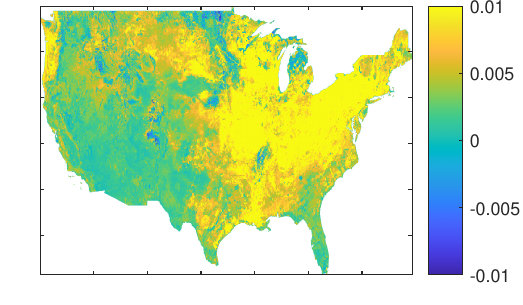}}
    \caption{The sensitivity of Solar-induced Fluorescence. Both radiation and root-zone soil moisture directly influence the rate of photosynthesis, which in turn affects the SIF signal. }
    \label{fig:intersif}
\end{figure*}

\section{Experiments}
\label{sec:results}

\subsection{Experimental setup}

 \paragraph{Training details.} All experiments are conducted on a server equipped with multiple NVIDIA V100 GPUs, Intel Xeon(R) Platinum 8260 CPU, and 256GB memory. The code is implemented with Python 3.9 and PyTorch 1.10.0.
 
 In this study, we split the pixels by $5\times5$ pixel block to avoid similar neighboring pixels and randomly select $80\%$ blocks as training pixels and the remaining $20\%$ for testing. Each pixel in our dataset has 572 weeks of temporal features and drought indices. We divide these 572 weeks into multiple windows for training and analysis. Each window consists of 100 weeks (approximately 2 years) designated as the training period, followed by 26 weeks (approximately half a year) designated as the prediction period. We then slide this window forward by 26 weeks (half a year) at a time, creating a total of 18 overlapping windows. To mitigate the impact of missing values (NaN) in the dataset, we impute the yearly average value for each week to maintain seasonal trends. Training is skipped for any NaN values in drought indices. Additionally, before training, we normalize each predictor {and drought index} by dividing it by its maximum value, scaling all values to a range between 0 and 1.
 
 During the training, we sample a batch of pixels randomly and shuffle the order of these windows to sequentially update the model. We train the model over 30 epochs where an epoch is defined as each training pixel being visited and trained once. After filtering out ocean locations, where most variables are NaN, the number of effective training pixels totals 380,801. The test set comprises 93,220 effective pixels. We set the batch size to 32, and employ Adam optimizer with a learning rate of 1e-4. The mean absolute error is used as the loss function. For categorical variable land cover, the embedding dimension is set at 4. For static numeric variables, the MLP uses a hidden dimension of 10 and an output dimension of 16.  Temporal features are first processed through a linear layer with a dropout rate of 0.1, mapping the dimension from 14 to 48. Then three layers of Transformer encoders and two layers of decoders with dimensions of 256 and 2 attention heads are used to learn from the projected temporal features. The model has 1M trainable parameters. Training on the entire training set takes approximately 80 minutes per epoch, and inference on the test set takes around 5 minutes.

 \paragraph{Baselines.}
 We consider state-of-the-art deep learning methods for time-series forecasting as baselines to evaluate our method. Note that these methods are mostly designed for time-series features without considering static features, including Transformer \cite{vaswani2017attention}, Informer \cite{zhou2021informer}, PatchTST \cite{nie2022time}, DLinear \cite{zeng2023transformers}, iTransformer \cite{liu2023itransformer}, TimesNet \cite{wu2022timesnet}, {and LSTM \cite{hochreiter1997long}}. We introduce the details of each baseline in Appendix 6.1.
 
\subsection{Results}

\paragraph{Performance Comparison.} We first compare \textit{SPDrought} with five widely recognized time-series forecasting baseline models. Table \ref{tab:baselines} presents the average mean absolute error across three runs on DroughtSet. The results demonstrate that \textit{SPDrought} has superior forecasting performance at forecasting 26 weeks of three drought indices at test locations compared with baselines. This outcome underscores its effectiveness in capturing the dynamics of the variables under study—Soil Moisture, Evaporative Stress Index, and Solar-induced Chlorophyll Fluorescence. 

Among the baselines, \textit{DLinear} has previously shown robust performance, outperforming several transformer-based methods \cite{zeng2023transformers} in forecasting. \textit{DLinear} decomposes the time series and uses two linear layers for trend and abnormality respectively. However, \textit{DLinear} encounters challenges with drought indices forecasting tasks because it uses prediction variables (drought indices) independently rather than leveraging all predictors and indices together. In contrast, transformer-based methods typically account for patterns among variables, resulting in better performance in our tasks.  It highlights the importance of learning the interplay of predictors to improve forecasting performance in drought prediction challenges. We also compare our model with a vanilla \textit{Transformer}. Except for the main difference in using static features, the \textit{Transformer} baseline uses embedding for temporal input tokenization as same as other methods, while our model considers the linear transformation for the temporal features to learn the representation across predictors. 

\paragraph{Drought Interpretation.} To examine the relative contribution of features to drought, we study the S2S drought in July 2012 in the CONUS. Specifically, we evaluate the influence of each variable on soil moisture prediction by comparing the integrated gradient value in Figure \ref{fig:intersm}. We select and present the top 3 variables showing their spatial patterns of the largest lag-1 integrated gradient during a drought week. Surface pressure, radiation, and PET show strong negative contributions ~(negative gradients) to soil moisture. That is, higher values of these metrics lead to decreases in soil moisture, thus potentially contributing to drought. Mechanistically, a higher surface pressure typically leads to drier weather by reducing the likelihood of rainfall, which in turn leads to drier soil conditions. In contrast, low-pressure systems are often associated with increased cloud cover and precipitation, which tend to enhance soil moisture \cite{bonan2019climate}. Additionally, both high solar radiation and high PET could enhance evapotranspiration, thus reducing soil moisture. Thus, the data-driven integrated gradient is able to recover the mechanistic dependency of soil moisture on climate conditions.  Figure \ref{fig:interesi} shows the top 3 predictors influencing the Evaporative Stress Index, i.e., radiation, surface pressure, and SIF, where higher levels of surface pressure contribute to increased evaporative stress index. Figure \ref{fig:intersif} shows pressure, radiation, and root zone soil moisture largely influence SIF. In particular, low radiation and low root zone moisture would reduce SIF, contributing to ecological drought. The impacts are especially apparent in the eastern US where vegetation is relatively denser compared to the western US. These observed results are consistent with first-order hydrological and ecological principles. The results on interoperability reveal the relative importance of these predictors to each drought index. The relative magnitudes and their spatial patterns contribute to discipline-specific understanding of the development and propagation of droughts.

\begin{table*}[h!]
\centering
\small
\caption{ Evaluation of drought prediction by soil moisture. The standard deviations are reported in the Appendix.}
\label{tab:acc_and_pre}

\setlength{\tabcolsep}{1mm}
\begin{tabular}{cccccccccc}
\hline
 & \textbf{SPDrought} & \textbf{Transformer}  & \textbf{Informer}  & \textbf{PatchTST} & \textbf{DLinear} & \textbf{iTransformer} & \textbf{TimesNet} & \textbf{LSTM}  \\
\hline
\textbf{Accuracy} & $\textbf{86.26}$     & $76.09$  & $72.16$ & $62.24$ & $62.85$ &  $77.18$  & $81.54$ & $77.45$ \\
\textbf{Precision} &  $\textbf{76.80}$    & $59.94$ & $53.40$  & $36.94$  & $37.96$ & $61.74$ & $68.98$ & $62.18$ \\
\hline
\end{tabular}
\end{table*}

\paragraph{Assessment of Drought Using Soil Moisture Percentiles.}

In this section, we use soil moisture as an example to assess drought by employing a percentile-based approach \cite{wang2018two}. Each data point of weekly soil moisture is compared against a multi-year average for the same calendar week, derived from historical data to represent typical moisture levels. We calculate the deviation of current soil moisture levels from these averages. We then use the 30th percentile as the threshold in our analysis. Soil moisture values below this percentile are considered as soil moisture drought. \textit{SPDrought} is compared with other methods in terms of accuracy and precision. Results on the test set are reported in Table \ref{tab:acc_and_pre}.

\begin{figure}[htbp]
    \centering
       \subfigure[Soil moisture drought detected from reanalysis data]{\includegraphics[width=0.45\linewidth]{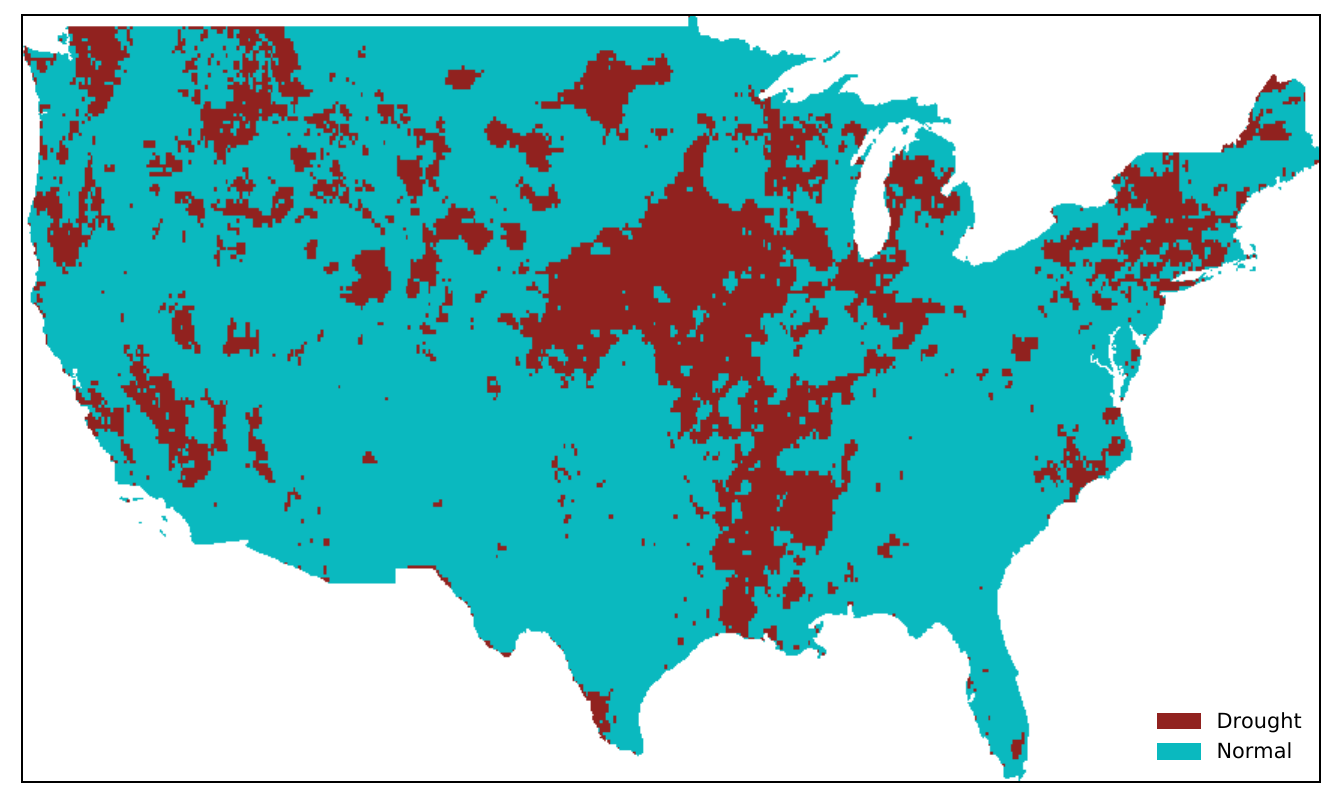}}
       \hspace{0.1cm}
       \subfigure[Soil moisture drought based on the prediction]{\includegraphics[width=0.45\linewidth]{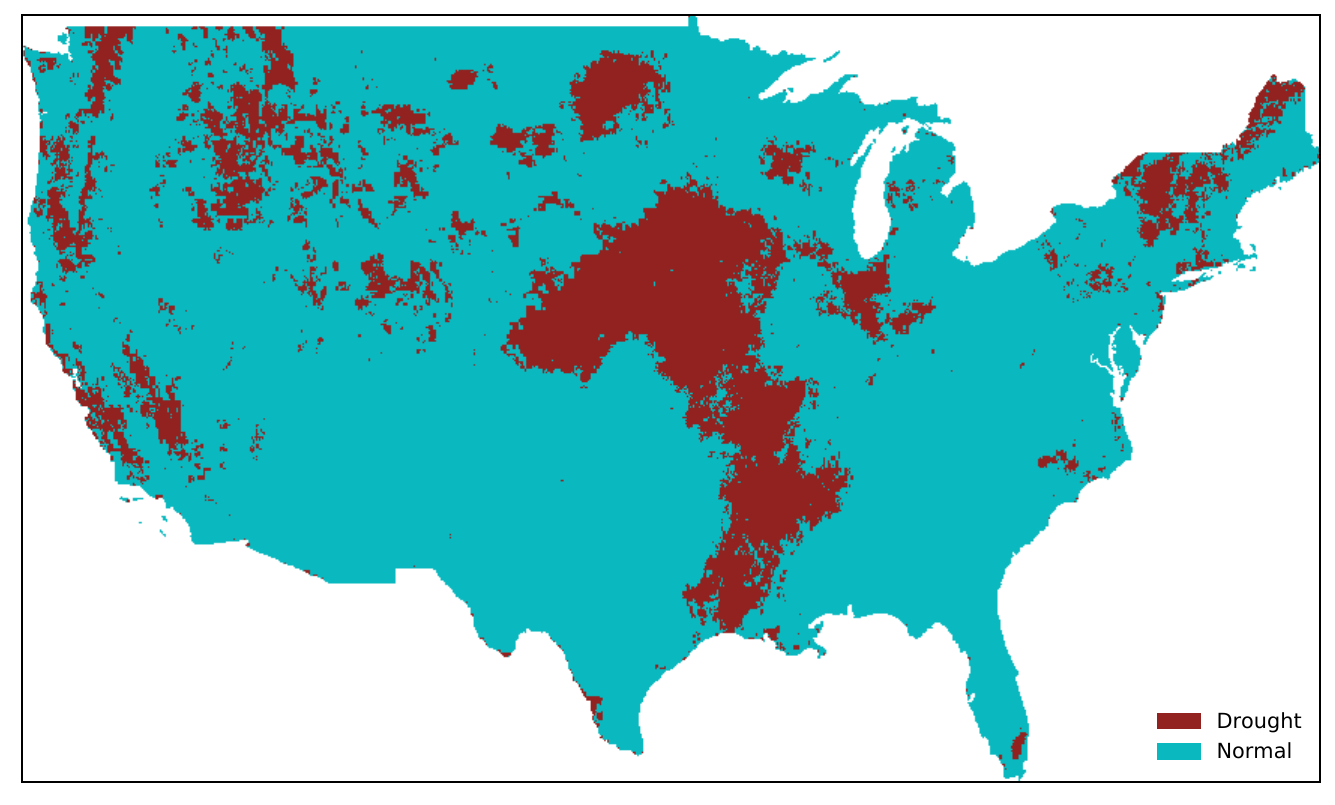}}
    \caption{Comparison of observation-based soil moisture drought and that predicted using our model \textit{SPDrought} with a lead time of 6 weeks in July 2012.}
    \label{fig:drought_pred}
\end{figure}

We visualize the predicted soil moisture drought in July 2012. Figure \ref{fig:drought_pred}.a. shows the observation-based soil moisture drought derived from the reanalysis product. For comparison, Figure \ref{fig:drought_pred}.b shows the drought derived from the soil moisture predicted using \textit{SPDrought}. The results show that \textit{SPDrought} successfully predicts the spatial pattern of observed soil moisture drought, especially in the Central Plains. In addition, we conduct ablation studies on static variables and model components, which are reported in Appendix 6.3.

\section{Conclusion}

This paper introduces DroughtSet, a specialized time-series forecasting dataset for predicting drought indices. It integrates vegetation and climate predictors that include both static and dynamic features. Based on DroughtSet, we propose \textit{SPDrought}, which exploits spatial-temporal interactions among the features to predict drought indices while providing interpretations on the impacts of each predictor on drought indices. Our findings contribute to improved understanding and prediction of drought development and propagation.

\noindent\textbf{Limitation and Future Work.}
This paper focuses on CONUS as a case study. Therefore, the trained model on DroughtSet is not suitable for direct deployment in other regions because of geographical differences. However, our method is not limited to CONUS and is expected to be effective in other regions, provided that relevant remote sensing and reanalysis datasets are available at a global scale. We suggest that expert knowledge is still needed to interpret the physical processes with complex mechanisms, such as the drivers of evaporative stress. In this study, we primarily examine the spatial influence of physical and climate conditions and vegetation dynamics on drought indices. Future case studies could benefit from comprehensive analyses of the temporal interplay among dynamic predictors and the dependencies related to S2S droughts.

\section*{Acknowledgements}
This material is based upon work supported by the U.S. National Science Foundation under award IIS-2202699 and IIS-2416895, by OSU President's Research Excellence Accelerator Grant, and grants from the Ohio State University's Translational Data Analytics Institute and College of Engineering Strategic Research Initiative.

\bibliography{aaai}

\clearpage
\section{Appendix}

\subsection{Baselines}
\label{app:baselines}

We ensured consistent representation dimensions and batch sizes across all baseline methods and adjusted model layers to maintain comparable training times. Additionally, other method-specific hyperparameters are adjusted to improve performance and ensure similar training times across all methods.

\begin{itemize}[leftmargin=*,topsep=0.1em,itemsep=0.3em]
     \item Transformer \cite{vaswani2017attention}: A vanilla Transformer for the time-series forecasting task.
     \item Informer \cite{zhou2021informer}: Informer introduces the ProbSparse self-attention mechanism for efficiently capturing long-range dependencies in time series forecasting.
     \item PatchTST \cite{nie2022time}:  PatchTST applies patching techniques to time series data, enhancing the transformer architecture's performance in capturing temporal patterns.
     \item DLinear \cite{zeng2023transformers}: DLinear is a simplified linear model which succeeds many transformer-based models. 
     \item iTransformer \cite{liu2023itransformer}: iTransformer inverted the duties of the self-attention mechanism and the feed-forward network to achieve better performance.
     \item TimesNet \cite{wu2022timesnet}:  TimesNet considers intraperiod and interperiod variations in 2D space for time series analysis. 
     \item {LSTM \cite{hochreiter1997long}:  LSTM is widely used in drought prediction. We follow the same architecture as existing studies \cite{yu2021spatial,danandeh2023novel,khan2024development}, with a convolution neural network as the feature extractor and an LSTM network to forecast drought indices.}

 \end{itemize}

  We use the same representation dimensions and batch size for all baselines. To ensure comparable training times across different methods, we utilize a 3-layer encoder and 2-layer decoder for Transformer, PatchTST, and iTransformer, aligning with our approach. For TimesNet, we employ a 2-layer encoder and 1-layer decoder to maintain training durations similar to those of other baselines. 

\subsection{Related Work}
\label{sec:related}

\subsubsection{AI in Climate}

AI in climate science has received significant attention in recent years. These advancements have enabled researchers to enhance climate models, improve climate prediction accuracy, and gain insights into the dynamics of the Earth system.  As examples, AI has been applied to predict El Niño-Southern Oscillation (ENSO) \cite{ham2019deep}, Typhoon detection \cite{park2023long}, and climate data downscaling \cite{park2022downscaling}. 

Traditional climate models rely on process-based representations and numerical methods to simulate climate and land surface processes, which can be computationally intensive and limited by the resolution and accuracy of climate forcing data. In contrast, deep learning models excel in recognizing complex patterns in large datasets, offering a complementary approach to process-based methods. In particular, some deep learning models show strong performance in processing temporal features. For example, \cite{park2023long} successfully used a transformer model to predict typhoon trajectories without relying on reanalysis data. \cite{liang2023airformer} proposed AirFormer for nationwide air quality prediction in China. \cite{nguyen2023climax} proposed a foundation model to forecast key climate {variables}. These examples underscore the growing efficacy and application of machine learning in climate science.

\subsubsection{Drought Prediction}

Drought prediction is one of the important tasks in climate science. Traditional climate models for drought prediction, which rely on process-based models and historical data, often struggle with the chaotic nature of climate systems. For example, the current generation of Earth system models (ESMs) has large biases in predicting precipitation at a sub-seasonal scale and thus flash drought \citep{zhang2021evaluation,mouatadid2023adaptive}. The Global Ensemble Forecast System based on process-based models, which holds the potential to implement operationally flash drought forecast guidance, also exhibits large prediction errors\citep{mo2020prediction}. Thus, many studies have highlighted the effectiveness of data-driven models in predicting droughts and identifying their key indicators. With the ability to deal with multicollinearity and non-linear relations among predictive features, machine learning (ML) models were applied to predict flash drought from weeks to months, measured by hydrological, meteorological, and agricultural metrics. These methods include support vector machines, random forests, decision trees, etc. { For example, \citet{adede2019mixed} predicts vegetation condition index using a simple ANN model. It is also applied for agricultural drought prediction using satellite images and climate indices \citet{marj2011agricultural}. However, these models were built using traditional ML approaches and often require handcrafted feature engineering or cannot effectively learn the feature from the data. As a result, they cannot exploit complex intercorrelation among different features and usually show limited predictive power. To tackle this challenge, recent studies have leveraged more advanced deep learning methods for flash drought predictions, which can learn hierarchical feature representations automatically from data and often outperform traditional ML methods \cite{ferchichi2022forecasting,gyaneshwar2023contemporary}. For example, deep neural networks have been applied in drought prediction \cite{agana2017deep,prodhan2021deep,kaur2020deep}. Models specifically designed for time-series data, such as LSTM, have also been used for predicting natural drought index \cite{vo2023lstm}, and agricultural drought conditions \cite{lees2022deep}. \citet{dikshit2021interpretable} further combine LSTM with the convolution neural network to predict the meteorological drought index in Eastern Australia and use SHapley Additive exPlanations to understand model outputs. Similar CNN-LSTM combined models are also used in \citet{yu2021spatial}, \citet{danandeh2023novel} and \citet{khan2024development}, where  \citet{yu2021spatial} predicts vegetation Index, \citet{danandeh2023novel} consider meteorological drought forecasting and\citet{khan2024development}  focus on predicting hydrological drought. \citet{amanambu2022hydrological} further adapts Transformer \citep{vaswani2017attention} to accurately forecast hydrological drought in the Apalachicola River. 

Compared with other studies using limited features \cite{dikshit2021long,khan2020prediction}, we consider more extensive features including physical conditions, climate conditions, and vegetation dynamics based on the underlying mechanisms of drought development. In particular, we consider the interplay between physical drivers and vegetation dynamics and advance our understanding of how climate and vegetation features and their spatial-temporal interactions regulate droughts. Thus we could learn more comprehensive representation from both static and temporal data to improve the forecasting performance. In addition, our method forecasts three different types of drought through multi-task learning using the shared representation without the need for extra computations to train separate models. }

\subsubsection{Time-series Forecasting}

Time series forecasting has been extensively studied across various domains, including climate science \cite{liang2023airformer}, traffic \cite{jiang2023pdformer}, and healthcare \cite{kaushik2020ai}. The complex and dynamic nature of time series data makes forecasting a challenging task.  Depending on the forecasting length, time series tasks can be categorized into long-term and short-term forecasting. Additionally, based on data types, there are univariate, multivariate, and spatio-temporal forecasting. LSTM \cite{hochreiter1997long} has been widely used in many time-series forecasting tasks. Recently, due to the tremendous success of the Transformer in natural language processing and computer vision, it has also been widely adopted in time-series forecasting problems. Researchers have proposed many variants of Transformers, such as Informer \cite{zhou2021informer}, Autoformer \cite{wu2021autoformer}, FEDformer \cite{zhou2022fedformer}, PatchTST \cite{nie2022time}, iTransformer \cite{liu2023itransformer}. Even though some transformers are proven not effective as linear-based methods like \cite{zeng2023transformers} in some tasks. The ability of Transformer to model global dependencies still makes them a popular choice for time series problems.

\subsection{Addition Results}

\subsubsection{Variable Importance Comparison}
\label{app:importance}

\begin{figure}
    \centering
    \includegraphics[width=1\linewidth]{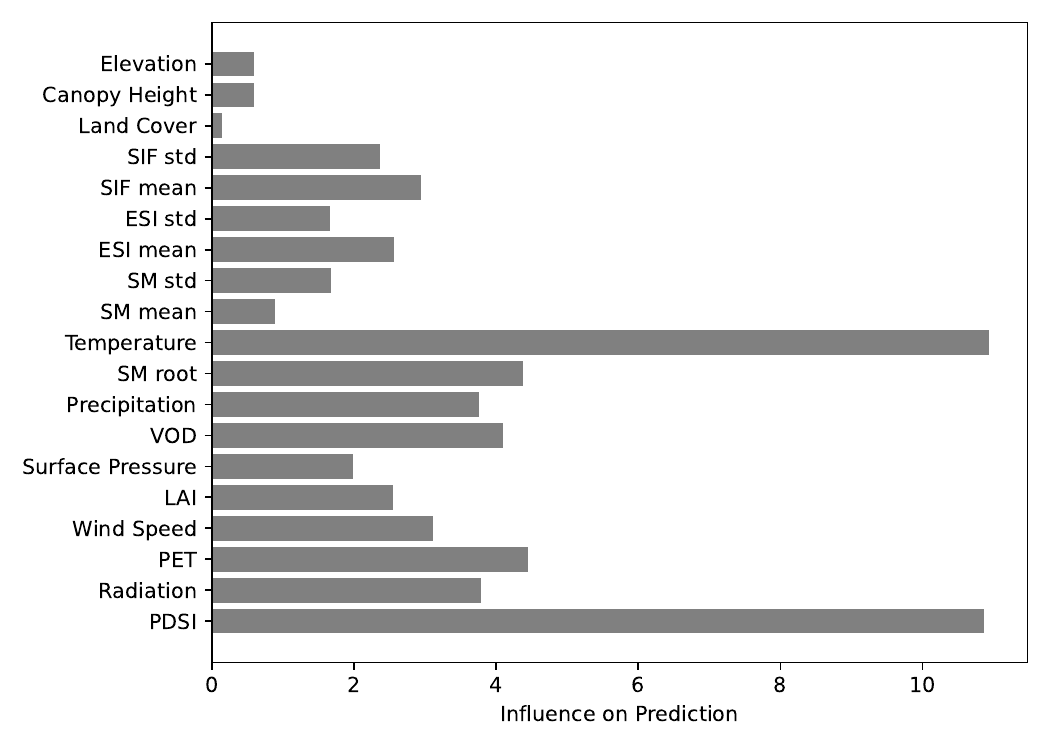}
    \caption{Relative Importance of Predictors}
    \label{fig:importance}
\end{figure}

In this section, we quantify the importance of each attribute to prediction from the perspective of prediction accuracy. We train models by excluding the predictive features one at a time and measure the change in model loss at the first epoch compared to a baseline model that uses all features. The differences shows in Figure \ref{fig:importance} illustrate the contribution of each feature to the performance of the models. The results also prove that static features such as vegetation dynamics and climate conditions are valuable in spatiotemporal forecasting for drought prediction tasks.

\subsubsection{Ablation Study on Static Features} 
\begin{figure}[H]
    \centering
    \subfigure{\label{ablation:sm}\includegraphics[width=0.45\linewidth, trim=10pt 4pt 8pt 10pt, clip]{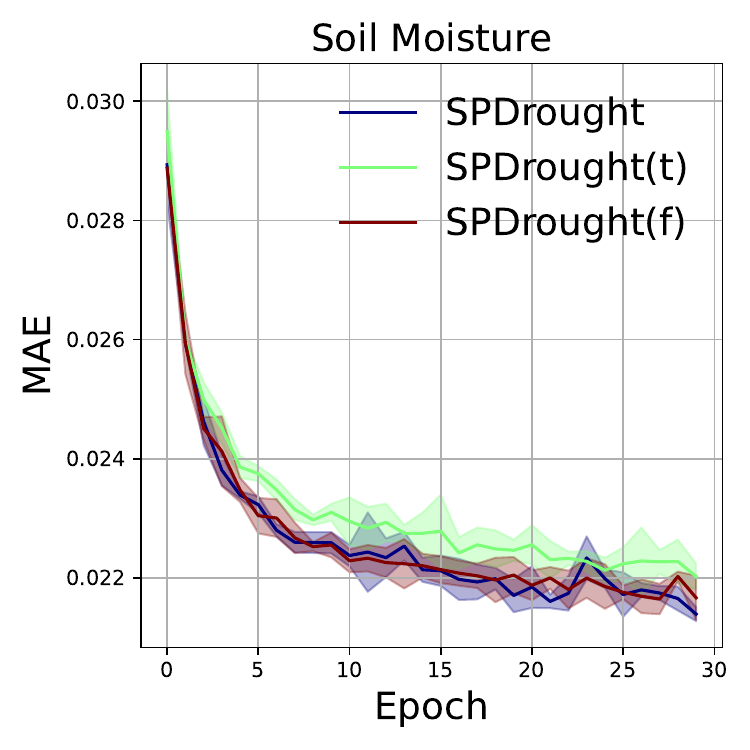}}
   \subfigure{\label{ablation:esi}\includegraphics [width=0.45\linewidth, trim=10pt 4pt 8pt 10pt, clip]{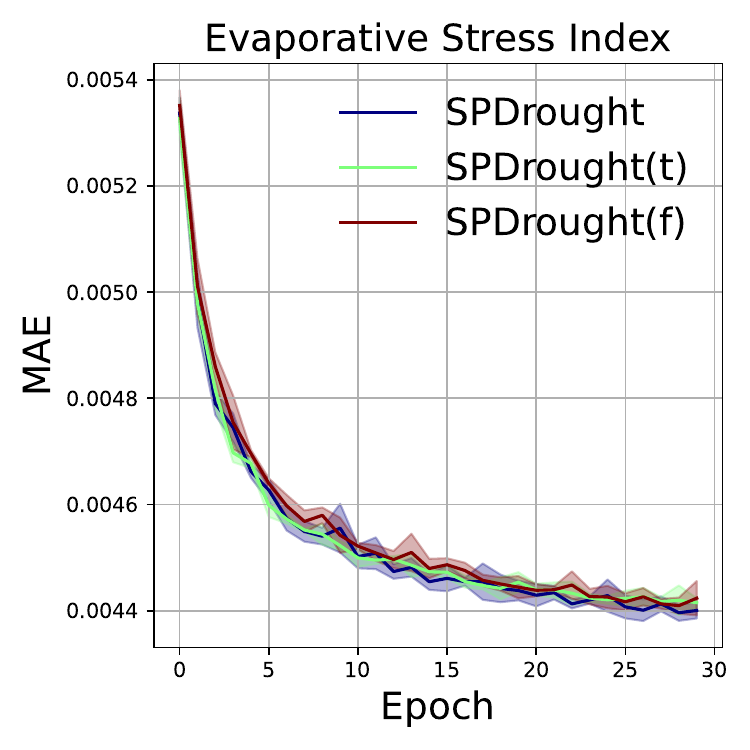}}
   
    \subfigure{\label{ablation:sif}\includegraphics[width=0.45\linewidth, trim=10pt 4pt 8pt 10pt, clip]{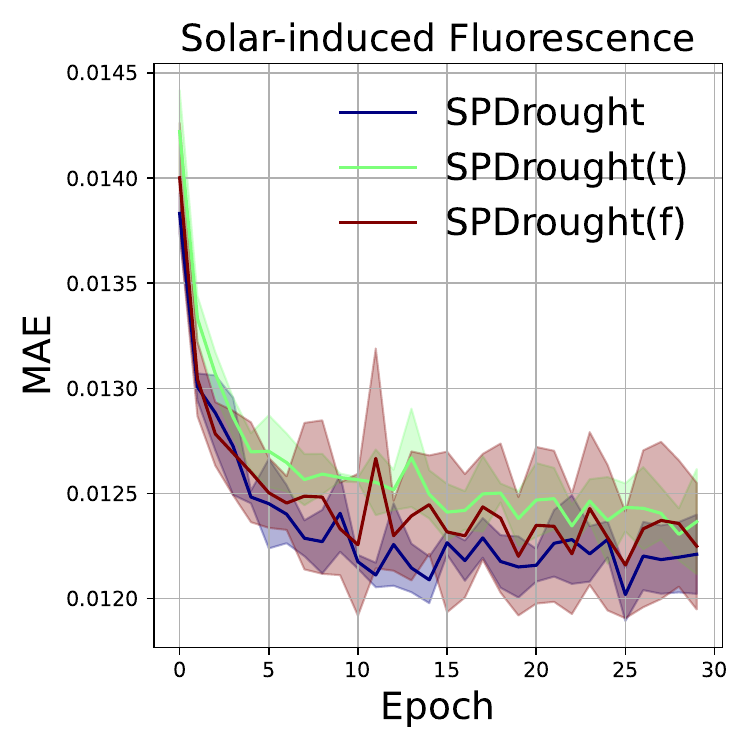}}
    \subfigure{\label{ablation:total}\includegraphics[width=0.45\linewidth, trim=10pt 4pt 8pt 10pt, clip]{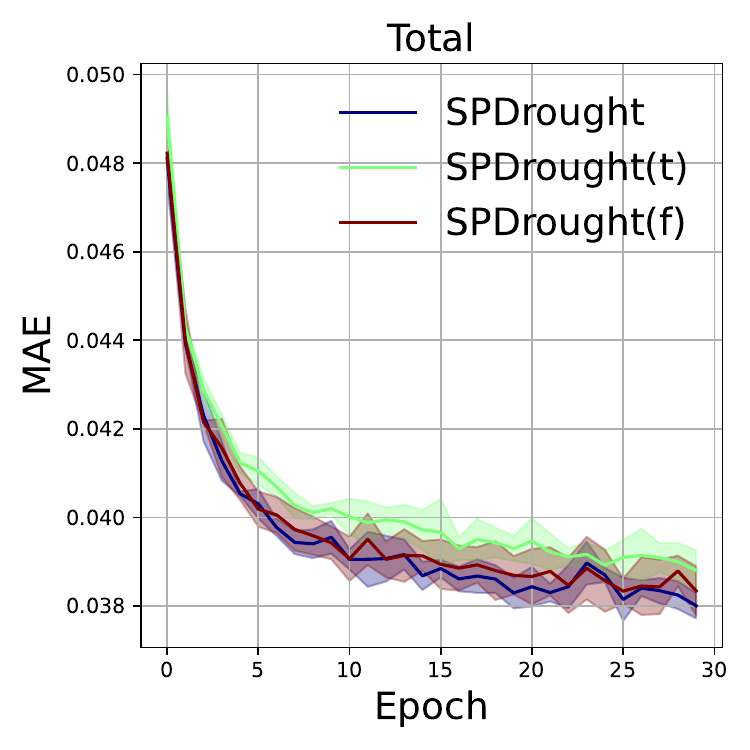}}
    \caption{Ablation study on static features and spatial-temporal fusion module. We run the experiment three times and report the average MAE loss on the test set.}
    \label{fig:ablation}
\end{figure}

In this section, we analyze the impact of each component within our model design. Initially, we assess the effectiveness of integrating static features into drought prediction tasks. To do this, we ablate the static features and compare \textit{SPDrought} with the modified version~(\textit{SPDrought}(t)) that does not utilize static features. \textit{SPDrought}(t) concatenates a full zero vector to the temporal representations before the transformer decoder and is also trained for 30 epochs. Then, we also investigate the impact of the spatial-temporal feature fusion module on model performance. So, we also conduct an additional experiment where we remove this module (denoted as \textit{SPDrought}(f)) and present results in Figure \ref{fig:ablation}. The results show that combining static predictors in representation can improve the forecasting performance across all drought prediction tasks, and the spatial-temporal feature fusion module consistently improves the performance and yields more stable outcomes. 

\begin{table*}[t!]
\centering
\caption{ Average MAE of the ablation study on multi-task learning}
\label{tab:abla_multitask}

\resizebox{\textwidth}{!}{
\begin{tabular}{cccccc}
\hline
\textbf{Model} & \textbf{Soil Moisture} & \textbf{Evaporative Stress Index} & \textbf{Solar-induced chlorophyll Fluorescence} & \textbf{Total} \\
\hline
\textbf{SPDrought(Single)} & $19.39_{\pm1.26}$
 &  $3.35_{\pm0.22}$ & $10.50_{\pm0.45}$  &  $33.24_{\pm0.93}$ \\
\textbf{SPDrought(Multi)} & $21.39_{\pm0.14}$ & $4.40_{\pm0.02}$ & $12.21_{\pm0.23}$ & $38.01_{\pm0.35}$ \\
\hline
\end{tabular}
}
\end{table*}

\begin{table*}
\centering
\caption{ Ablation study on model parameters}
\label{tab:abla_model_para}

\resizebox{\linewidth}{!}{
\begin{tabular}{cccccc}
\hline
 MAE ($\times 10^{-3}$)& \textbf{SPDrought} & \textbf{SPDrought(w/o Encoder)}  & \textbf{SPDrought(w/o Decoder)}  & \textbf{SPDrought(50)}  \\
\hline
\textbf{Soil Moisture} & $21.39_{\pm0.14}$     & $28.20_{\pm0.17}$  &  $23.87_{\pm0.33}$ & $22.46_{\pm0.14}$  \\
\textbf{Evaporative Stress Index} &  $4.40_{\pm0.02}$    & $5.51_{\pm0.01}$ & $4.72_{\pm0.04}$  & $4.51_{\pm0.01}$   \\
\textbf{Solar-induced chlorophyll Fluorescence} & $12.21_{\pm0.23}$  & $17.24_{\pm0.58}$ & $12.41_{\pm0.13}$ & $12.30_{\pm0.15}$    \\
\textbf{Total} & $38.01_{\pm0.35}$  & $50.95_{\pm0.70}$ & $41.01_{\pm0.48}$ &  $39.27_{\pm0.06}$   \\
\hline
\end{tabular}
}
\end{table*}

\begin{table*}[h!]
\centering
\renewcommand{\arraystretch}{1.1}
\caption{ Average MAE over three runs of experiments on drought indices over 26 weeks using the temporal split}
\label{tab:baselines_timesplit}
\resizebox{\textwidth}{!}{
\begin{tabular}{cccccccccc}
\hline
 MAE ($\times 10^{-3}$)& \textbf{SPDrought} & \textbf{Informer}  & \textbf{PatchTST} & \textbf{DLinear} & \textbf{iTransformer} & \textbf{TimesNet} & {\textbf{LSTM}} \\
\hline
\textbf{Soil Moisture} & $47.77_{\pm0.58}$    &  $51.97_{\pm0.09}$ & $48.27_{\pm0.28}$ & $48.66_{\pm0.81}$ & $49.55_{\pm0.12}$  & $48.48_{\pm0.21}$ & $53.61_{\pm0.16}$\\
\textbf{Evaporative Stress Index} & $5.94_{\pm0.02}$     & $6.37_{\pm0.06}$  & $6.52_{\pm0.04}$  & $6.69_{\pm0.10}$  & $6.39_{\pm0.03}$ & $6.20_{\pm0.03}$ &  $6.48_{\pm0.04}$ \\
\textbf{Solar-induced chlorophyll Fluorescence} & $22.26_{\pm0.05}$   & $25.91_{\pm0.18}$ & $29.16_{\pm0.47}$  & $29.28_{\pm0.38}$ & $25.28_{\pm0.15}$ & $25.27_{\pm0.59}$ & $26.04_{\pm0.20}$ \\
\textbf{Total} & $75.94_{\pm0.65}$  & $84.24_{\pm0.19}$ & $83.93_{\pm0.78}$  & $84.60_{\pm1.23}$ & $81.20_{\pm0.13}$    & $79.93_{\pm0.81}$ &  $86.10_{\pm0.23}$ \\
\hline
\end{tabular}
}
\end{table*}

\subsubsection{Ablation Study on Multi-task Training}

Here, we compare multi-task training with single-task training and report the results in Table \ref{tab:abla_multitask}. The ablation study highlights the efficiency of the multi-task learning approach. While training the SPDrought model on multiple tasks simultaneously (\textit{SPDrought(Multi)}), it achieves comparable results to training on individual tasks (\textit{SPDrought(Single)}) but in about one-third the time. This demonstrates that multi-task learning can significantly speed up the training process without a substantial drop in accuracy, making it a valuable strategy when time and computational resources are limited.

\subsubsection{Ablation Study on Model Parameters}

In this section, we explore the contribution of each component by removing the component from \textit{SPDrought}. We conduct this study by creating variants of the model: \textit{SPDrought} without the Transformer Encoder (\textit{SPDrought(w/o Encoder)}), \textit{SPDrought} without the Transformer Decoder (\textit{SPDrought(w/o Decoder)}), and \textit{SPDrought} with a reduced training window of 50 weeks, approximately one year (\textit{SPDrought(50)}). These variants help us understand the role of each component in capturing temporal dependencies and learning drought patterns from historical data. As shown in Table \ref{tab:abla_model_para}, the Transformer encoder effectively helps to capture time dependence and thus significantly improves the overall performance. Furthermore, reducing the training window to 50 weeks (\textit{SPDrought(50)}) slightly affects the model's accuracy. It suggests that SPDrought can effectively learn drought patterns even with limited historical data. However, extended historical data contributes to better model performance, highlighting the importance of a more comprehensive dataset for training.

\subsubsection{Comparison on Temporal Splitting}

In the previous comparison, we evaluate each method on test pixel regions. Here, we adopt a temporal split in the data, assessing the baseline methods and our approach for predicting drought indices over the next 26 weeks which are not seen during training.

\subsubsection{Supplementary Results}

\begin{table*}[h!]
\centering
\caption{ Evaluation of drought prediction by soil moisture with standard deviations. }
\label{tab:acc_and_pre_full}

\resizebox{\linewidth}{!}{
\begin{tabular}{cccccccccc}
\hline
 & \textbf{SPDrought} & \textbf{Transformer}  & \textbf{Informer}  & \textbf{PatchTST} & \textbf{DLinear} & \textbf{iTransformer} & \textbf{TimesNet} & \textbf{LSTM}  \\
\hline
\textbf{Accuracy} & $86.26_{\pm0.11}$     & $76.09_{\pm0.11}$  & $72.16_{\pm0.34}$ & $62.24_{\pm4.63}$ & $62.85_{\pm0.01}$ &  $77.18_{\pm0.04}$  & $81.54_{\pm0.91}$ & $77.45_{\pm0.26}$\\
\textbf{Precision} &  $76.80_{\pm0.17}$    & $59.94_{\pm0.19}$ & $53.40_{\pm0.57}$  & $36.94_{\pm7.70}$  & $37.96_{\pm0.02}$ & $61.74_{\pm0.07}$ & $68.98_{\pm1.51}$ & $62.18_{\pm0.43}$ \\
\textbf{p-value} &  -    & $1.49 \times 10^{-4}$ & $9.96 \times 10^{-5}$  & $1.17 \times 10^{-2}$  & $6.78 \times 10^{-6}$ &  $1.68 \times 10^{-5}$ & $9.77 \times 10^{-3}$ &  $5.57 \times 10^{-4}$  \\
\hline
\end{tabular}
}
\end{table*}

We present the Table \ref{tab:baselines} with standard deviations in the Table \ref{tab:baselines_std}. We also include the accuracy and precision with standard deviations in Table \ref{tab:acc_and_pre_full}, where we report the significance of improvements using paired t-tests on precision.

\begin{table*}[h!]
\centering
\caption{ Average mean absolute error over three runs of experiments with standard deviations}
\label{tab:baselines_std}
\resizebox{\linewidth}{!}{
\begin{tabular}{cccccccccc}
\hline
 MAE ($\times 10^{-3}$)& \textbf{SPDrought} & \textbf{Transformer}  & \textbf{Informer}  & \textbf{PatchTST} & \textbf{DLinear} & \textbf{iTransformer} & \textbf{TimesNet} & {\textbf{LSTM}} \\
\hline
\textbf{Soil Moisture} & $\textbf{21.39}_{\pm0.14}$     & $34.56_{\pm0.24}$  & $38.08_{\pm0.14}$ & $36.32_{\pm0.19}$ & $47.61_{\pm0.04}$ & $32.34_{\pm0.09}$  & $25.96_{\pm0.46}$ & $31.36_{\pm0.40}$ \\
\textbf{ESI} &  $\textbf{4.40}_{\pm0.02}$    & $5.99_{\pm0.07}$ & $6.37_{\pm0.06}$  & $6.37_{\pm0.00}$  & $6.82_{\pm0.01}$  & $6.06_{\pm0.01}$ & $5.11_{\pm0.02}$ &  $5.83_{\pm0.09}$ \\

\textbf{SIF} & $\textbf{12.21}_{\pm0.23}$  & $16.00_{\pm0.23}$ & $17.71_{\pm0.33}$ & $21.36_{\pm0.19}$  & $20.99_{\pm0.03}$ & $15.47_{\pm0.07}$ & $14.11_{\pm0.03}$ & $15.35_{\pm0.02}$ \\
\textbf{Total} & $\textbf{38.01}_{\pm0.35}$  & $56.56_{\pm0.05}$ &  $62.16_{\pm0.43}$ & $64.05_{\pm0.34}$  & $75.41_{\pm0.05}$ & $53.87_{\pm0.16}$    & $45.18_{\pm0.44}$ &  $52.54_{\pm0.40}$ \\
\hline
\end{tabular}
}
\end{table*}

\subsection{Data Source}
\label{app:source}

We collect data from the following source:
\begin{itemize}
    \item NLDAS \cite{xia2012continental}: NLDAS is provided by NASA, collected from \url{https://ldas.gsfc.nasa.gov/nldas}.
    \item SMAP \cite{das2018smap}: SMAP is a public dataset provided by NASA, collected from  \url{https://smap.jpl.nasa.gov/}.
    \item ALEXI ET was obtained upon request from Thomas R. Holmes and Christopher R. Hain on 28 January 2020 and was presented in \cite{holmes2018microwave, liu2021global}.
    \item CSIF \cite{zhang2018global}: CSIF dataset is under CC BY 4.0,  collected from \url{https://figshare.com/articles/dataset/CSIF/6387494}.
    \item ERA5 \cite{munoz2021era5}: ERA5 is provided by the European Centre for Medium-Range Weather Forecasts under Copernicus license, data collected from \url{https://www.ecmwf.int/en/forecasts/dataset/ecmwf-reanalysis-v5}.
    \item SRTM \cite{nasa_jpl_2013}: NASA Shuttle Radar Topography Mission (SRTM) datasets are provided under the U.S. Geological Survey (USGS),  collected from \url{https://lpdaac.usgs.gov/products/srtmgl1v003/}.
    \item VODCA \cite{moesinger2020global}: VODCA is under CC BY 4.0, colleted from 
    \url{https://zenodo.org/records/2575599}.
    \item MODIS \cite{myneni2015mcd15a3h}: MODIS is provided by NASA,  collected from \url{https://modis.gsfc.nasa.gov/}.
    \item GLAD \cite{potapov2021mapping}: Global Land Analysis \& Discovery (CC BY),  collected from \url{https://glad.umd.edu/}.
    \item NLCD \cite{homer2012national}: Nation Land Cover Database is in the public domain, provided under USGS,  collected from \url{https://www.usgs.gov/centers/eros/science/national-land-cover-database}.
\end{itemize}

\end{document}